\documentclass{article} 
\usepackage{iclr2026_conference,times}

\usepackage{amsmath,amsfonts,bm}

\def\eqref#1{equation~\ref{#1}}

\def\1{\bm{1}}

\DeclareMathAlphabet{\mathsfit}{\encodingdefault}{\sfdefault}{m}{sl}
\SetMathAlphabet{\mathsfit}{bold}{\encodingdefault}{\sfdefault}{bx}{n}

\usepackage{booktabs}
\usepackage[hidelinks]{hyperref}
\usepackage{url}
\usepackage{graphicx} 
\usepackage{csquotes}
\usepackage{siunitx}
\usepackage{capt-of}
\usepackage{needspace}
\usepackage{orcidlink}
\usepackage{float}
\title{Is More Data Worth the Cost? Dataset Scaling Laws in a Tiny Attention-Only Decoder}

\author{
\renewcommand{\thefootnote}{\Alph{footnote}}
\begin{tabular}{l l l}
\textbf{Götz-Henrik Wiegand$^{\diamond}$\,\orcidlink{0009-0009-0392-056X}$^{A}$} &
\textbf{Lorena Raichle$^{\diamond*}$\,\orcidlink{0009-0000-4242-2728}$^{B}$} &
\textbf{Rico Städeli$^{\diamond}$\,\orcidlink{0009-0007-9741-6633}$^{C}$} \\
\textbf{Tomas Hrycej$^{\diamond}$\,\orcidlink{0009-0008-1066-5045}$^{D}$} &
\textbf{Bernhard Bermeitinger$^{\triangleright}$\,\orcidlink{0000-0002-2524-1850}$^{E}$} &
\textbf{Siegfried Handschuh$^{\diamond}$\,\orcidlink{0000-0002-6195-9034}$^{F}$}
\end{tabular}
\\[0.5em]
\normalsize
\centerline{{\tiny$^{\diamond}$}\small Institute of Computer Science, University of St.\ Gallen, St.\ Gallen, Switzerland}\\
\centerline{{\tiny$^{\triangleright}$}\small Institute of Computer Science in Vorarlberg, University of St.\ Gallen, Dornbirn, Austria}\\
[0.5em]
\centerline{\texttt{\ \{name.lastname\}@unisg.ch}; \texttt{$^{*}$\{name.lastname\}@student.unisg.ch}}\\
}

\iclrfinalcopy 
\begin{document}

\maketitle
\begingroup
\renewcommand{\thefootnote}{}
\footnotetext{
\tiny
\fontsize{5}{6}\selectfont
\orcidlink{0009-0009-0392-056X}$^{A}$: 0009-0009-0392-056X \quad
\orcidlink{0009-0000-4242-2728}$^{B}$: 0009-0000-4242-2728 \quad
\orcidlink{0009-0007-9741-6633}$^{C}$: 0009-0007-9741-6633 \quad
\orcidlink{0009-0008-1066-5045}$^{D}$: 0009-0008-1066-5045 \quad
\orcidlink{0000-0002-2524-1850}$^{E}$: 0000-0002-2524-1850 \quad
\orcidlink{0000-0002-6195-9034}$^{F}$: 0000-0002-6195-9034
}
\endgroup

\begin{abstract}
Training Transformer language models is expensive, as performance typically improves with increasing dataset size and computational budget. Although scaling laws describe this trend at large scale, their implications in controlled, smaller-scale settings remain less explored.
In this work, we isolate dataset-size effects using a strongly reduced \emph{attention-only} decoder architecture. By training on progressively larger power-of-two subsets, we observe smooth performance improvements accompanied by clear diminishing returns, consistent with scaling-law behavior. Using only about 30\% of the training data is sufficient to reach approximately 90\% of the full-data validation token-level accuracy.
These results provide actionable insights into dataset scaling in a controlled, component-isolated setting and offer practical guidance for balancing dataset size and computational cost in compute- and data-restricted environments, such as small research labs and exploratory model development.

\end{abstract}

\section{Introduction}
Increasing model and dataset size reliably improves Large Language Model (LLM) performance but incurs substantial computational and energy costs. In practice, pretraining often involves difficult trade-offs between training cost and performance, making it crucial to understand when additional data yields diminishing returns and how much is sufficient to approach near-saturated performance. 

Prior work on scaling laws has shown that LLM performance improves smoothly with increasing model size, data, and compute, following predictable power-law relationships~\citep{kaplanScalingLawsNeural2020a, hoffmannTrainingComputeOptimalLarge2022a}. Most existing scaling law studies are conducted at industrial scale, where these factors are scaled jointly, making it difficult to disentangle their individual contributions and to attribute observed performance trends to specific model components. Throughout this work, \emph{scaling laws} refer to the formulation introduced by~\citet{kaplanScalingLawsNeural2020a}. Their analysis spans model sizes from tens of thousands to billions of parameters and wide ranges of dataset size. At the same time, recent advances(e.g. \emph{NanoGPT}\footnote{\tiny Beating GPT-2 for under \$100: the nanochat journey by Andrej Karpathy: https://github.com/karpathy/nanochat/discussions/481}) have shown that even comparatively small and reduced-capacity models can exhibit strong capabilities, raising the question of whether the behavior observed at industrial scale persists in tiny models. Resolving this question through the controlled analyses presented in this work is particularly valuable for compute- and data-constrained pretraining settings, where identifying diminishing returns in dataset scaling is essential for reducing training cost.

We adopt a controlled experimental setup that isolates dataset-size effects under a fixed-capacity \emph{attention-only} decoder architecture. Our objective is to determine whether increasing training data reproduces the characteristic diminishing returns predicted by scaling laws, when learning is primarily restricted to the self-attention mechanism. This design emphasizes interpretability and controlled analysis over absolute model performance. We train our model on progressively larger power-of-two training subsets and analyze how dataset size and the token-to-parameter ratio influence training dynamics and final performance.\\

Our main contributions are:
\begin{itemize}
    \item A fixed-capacity \textbf{attention-only architecture} that freezes pretrained embeddings and the output projection while removing Multi-Layer Perceptron (MLP) sublayers.
    \item An analysis of \textbf{dataset representativeness and robustness} via token-distribution and multi-seed experiments.
    \item A controlled \textbf{dataset-scaling study} conducted with a fixed \textbf{attention-only decoder} architecture.

    \item A characterization of the \textbf{cost--performance trade-off} in dataset scaling, showing that moderate dataset sizes yield most full-data performance at reduced training cost.
\end{itemize}

Taken together, our findings offer practical guidance on data-efficient pretraining, with particular relevance for compute-constrained or exploratory development of small and tiny models.

\section{Related Work}

\subsection{Scaling Laws and Diminishing Returns}

The relationship between LLM performance and scale was formalized by ~\citet{kaplanScalingLawsNeural2020a}, who showed that validation loss follows a
power-law decay as model size, dataset size, and compute increase. Within this framework,
performance is governed by the relationship between the number of training tokens
and the number of trainable model parameters. A key implication of these scaling
laws is the presence of diminishing returns: each doubling of data or parameters
yields progressively smaller performance improvements. Together, these findings
established scaling laws as a predictive framework for large-scale language model
training.

Subsequent work refined and extended these insights. ~\citet{hoffmannTrainingComputeOptimalLarge2022a}
showed that many LLMs, including GPT-3 and Gopher, were under-trained
relative to their parameter count. They introduced a compute-optimal scaling prescription \enquote{Chinchilla}, which emphasizes balancing model size and dataset size
under a fixed compute budget. This perspective shifts emphasis from scaling model size to
understanding data efficiency under a fixed compute budget, motivating our focus
on dataset scaling. Other studies explored the limits of data scaling under constrained settings. ~\citet{hernandezScalingLawsInterpretability2022a} examined the effect of repeated exposure to the same
dataset and found that excessive repetition can lead to overfitting and degraded
performance, indicating that more training steps or epochs do not always yield
better generalization. Similarly, ~\citet{chenRevisitingScalingLaws2025} identified a
``sub-scaling'' regime in which performance gains slow due to redundancy in the
training data. These findings reinforce the notion that dataset size alone does
not linearly translate into improved performance. ~\citet{muennighoffScalingDataConstrainedLanguage2025} further investigated data efficiency by studying
training under data-constrained regimes. They showed that repeating data can remain useful for a limited number of epochs, after which returns diminish sharply. This observation informs
our choice of training budgets and epoch counts in subset-based experiments.

\subsection{Subset Pretraining and Data Efficiency}

Complementary to theoretical scaling-law analyses, recent work has explored practical strategies for improving data efficiency by training on subsets, which correspond to fractions of large datasets. Our work also builds on recent advances in efficient transformer training via subset pretraining. ~\citet{sporerEfficientNeuralNetwork2024} proposed \textit{subset pretraining}, showing that training on small subsets as a pretraining strategy before switching to full dataset training can match the performance of full-dataset training at a fraction of the computational cost in computer vision tasks. Their findings motivate our investigation of whether similarly small and representative subsets can capture most of the learning signal in decoder-only LLMs.

~\citet{wangFarewellAimlessLargescale2023} introduced Influential Subset Selection (ISS), demonstrating that carefully selected, extremely small subsets can achieve
competitive pretraining performance and consistently outperform randomly
sampled or heuristic subset baselines. These results highlight the importance of
data selection and representativeness, reinforcing the view that model
performance depends not only on data volume, but also on the quality and
informativeness of the training data.

Taken together, scaling laws characterize how performance improves with increasing data and
compute, while subset pretraining methods demonstrate that much of this
performance can be achieved using only a fraction of the available data.
Prior work makes it difficult to attribute observed behavior
to specific architectural components. In contrast, our work connects these two
lines of research by studying dataset scaling under a fixed decoder-only
architecture and optimization procedure, while explicitly isolating the
self-attention mechanism. 

\subsection{Architecture Reductions}
The \emph{attention-only} architecture used in this study is based on prior work by~\citet{bermeitingerReducingTransformerArchitecture2024}, which showed that MLP sublayers in the decoder can be removed without significant performance degradation.

\section{Architecture}\label{sec:architecture}
\subsection{Base Model}
\begin{figure*}[htb]
    \centering
    \begin{minipage}[c]{0.48\textwidth}
        \centering
        \includegraphics[width=0.8\linewidth, keepaspectratio]{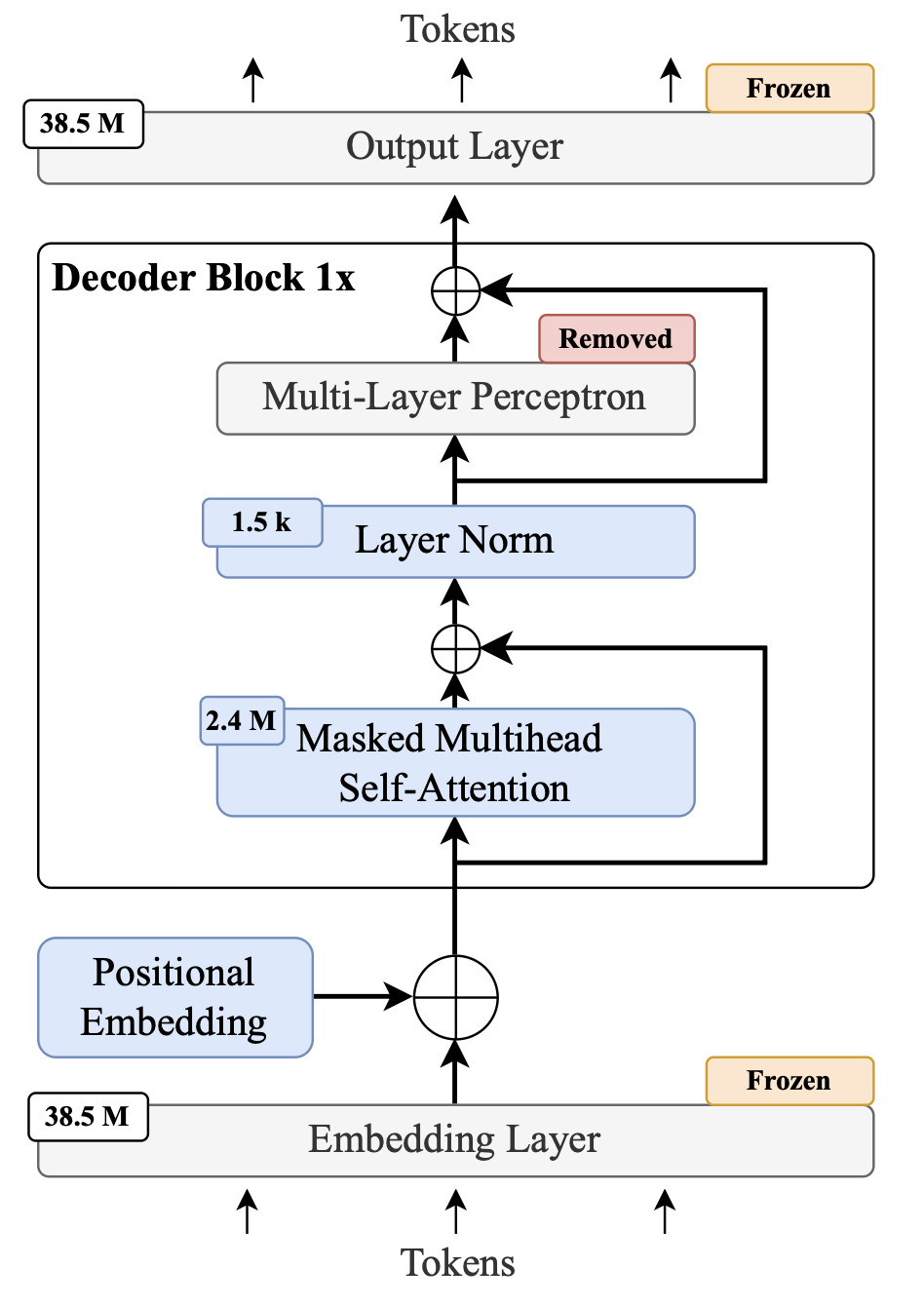}
        \caption{Architecture graph of the \emph{attention-only} architecture with individual layers.}
        \label{fig:decoder}
    \end{minipage}
    \hfill
    \begin{minipage}[c]{0.48\textwidth}
        \centering
        \captionof{table}{Key overview of the \emph{attention-only} architecture with insights to architecture reductions and the parameter count with the amount of total parameters and trainable parameters.}
        \vspace{0.2cm}
        \label{tab:model_config}
        
        \setlength{\tabcolsep}{4pt}
        \renewcommand{\arraystretch}{1.1}
        \small
        
        \begin{tabular}{ll}
         
            \multicolumn{2}{c}{\textbf{Architecture}} \\
            \midrule
            Architecture Type & attention-only architecture\\
            Sequence Length & 1024 \\
            Embedding Dimension & 768 \\
            Number of Attention Heads & 1 \\
            Number of Decoder Layers & 1 \\
            Remove MLP & True \\
            Frozen Embedding Layer & True \\
            Frozen Output Layer & True \\
            \midrule
            \multicolumn{2}{c}{\textbf{Parameter Count}} \\
            \midrule
            Total Parameters & 79,556,352 \\
            Trainable Parameters & 2,361,600 \\
        \end{tabular}
    \end{minipage}
\end{figure*}

To study the persistence of scaling-law behavior at small model scales, we adopt a decoder-only Transformer configuration based on the 117M GPT-2 design often also referred to as GPT-2 Base or GPT-2 Small~\citep{radfordLanguageModelsAre2019a}. We adopt a controlled setup that yields tiny, reduced-capacity models through two architectural constraints: freezing the embedding and output layers in Section~\ref{subsec:3.1.1} and removing the MLP sublayers in Section~\ref{subsec:3.1.2}. Under this configuration, only the parameters of the self-attention layers are updated during training. Token embeddings, positional embeddings, and the output projection layer are frozen, and the MLP sublayers of decoder blocks are removed entirely, as indicated in Fig.~\ref{fig:decoder}.

The 117M GPT-2 design~\citep{radfordLanguageModelsAre2019a} provides a standardized architecture together with a widely used tokenizer and pretrained embedding space at minimal scale. The availability of pretrained embeddings for such small models is limited in practice; GPT-2 thus offers a pragmatic and well-established basis for isolating dataset-scaling effects by leveraging its pretrained embedding layer, thereby avoiding confounding variability from representation learning. The model employs learned absolute positional embeddings added to token embeddings. Our design comprises approximately 2.4 million trainable parameters and is trained on datasets ranging from 0.1 million to 134 million tokens. This range spans regimes from strongly data-limited to token-to-parameter ratios suggested by commonly cited compute-optimal heuristics, while keeping model capacity constant~\citep{hoffmannTrainingComputeOptimalLarge2022a}. 
\subsubsection{Freezing Pretrained Embedding and Output Layer}\label{subsec:3.1.1}
In scaling studies, embedding parameters are often abstracted away from effective model capacity in scaling analyses~\citep{kaplanScalingLawsNeural2020a}, or explicitly frozen to prevent representation learning from dominating learning behavior~\citep{bermeitingerReducingTransformerArchitecture2024}. This consideration is especially critical in the tiny-model regime studied here. In our setting, the pretrained token embedding matrix constitutes a substantial fraction of the total parameter count(see ratio of total parameters to trainable parameters in Table~\ref{tab:model_config}) and would dominate the trainable capacity if left unfrozen, relative to the comparatively small number of parameters in the decoders. In preliminary experiments where the embedding layer was trained jointly with the decoder, we observed rapid overfitting and unstable training dynamics, with apparent performance gains driven primarily by embedding adaptation rather than by learning contextual relationships through self-attention. Training the embedding layer would fundamentally alter the learning problem across data regimes, since in small-data settings much of the model capacity would be spent relearning token representations. To avoid this imbalance, we freeze the pretrained token embedding layer and the output projection layer throughout training, leveraging GPT-2 representations. In GPT-2, the output projection is weight-tied to the input embedding matrix, such that next-token prediction compares the decoder hidden state against all token embeddings. Freezing the output layer therefore fixes this shared embedding space. This design choice stabilizes optimization in low-data regimes by preventing the model from relearning token-level representations from scratch. As a result, performance differences across subset sizes can be attributed directly to dataset scaling effects in the self-attention mechanism rather than to dataset-dependent representation learning.

\subsubsection{Attention-Only Transformer via MLP Removal}\label{subsec:3.1.2}
To further operate in the tiny-model regime, we remove the MLP sublayers from each Transformer decoder block, yielding an \emph{attention-only} architecture. While MLPs are commonly justified as the primary source of nonlinearity in Transformer models, recent work has shown that self-attention itself is highly nonlinear due to its data-dependent weighting and Softmax-based similarity computation, and can capture complex relationships even without an explicit feed-forward network. ~\citet{bermeitingerReducingTransformerArchitecture2024} demonstrate that attention-only architectures can achieve competitive performance while substantially reducing the number of trainable parameters. Since the MLP typically accounts for the majority of parameters in a decoder block, removing it reduces model capacity in a controlled and interpretable manner, allowing observed scaling behavior to be attributed more directly to the self-attention mechanism.

\subsection{Implications of Model Reduction}

Freezing pretrained components and removing the MLP sublayers
substantially reduce the number of trainable parameters. Table~\ref{tab:model_config} summarizes the final \textit{attention-only} model architecture used for all subset-based scaling experiments in this work. Under this configuration, the full model contains
79{,}556{,}352 parameters in total, of which only
2{,}361{,}600 are trainable.

As a consequence of freezing the embedding layer, the token embedding layer is initialized from pretrained 117M GPT-2 weights and excluded from optimization. GPT-2 does not define an explicit padding token. Since reusing pretrained 117M GPT-2 embeddings requires preserving the original vocabulary size, introducing an additional padding token would break compatibility with the pretrained embedding matrix. We therefore reuse the end-of-sequence (EOS) token for padding, mask these positions during self-attention, and exclude them from loss and accuracy computation by assigning the ignore label (e.g., \texttt{-100}) to padded targets.\\

\section{Dataset}\label{sec:dataset}
\subsection{Dataset Description}
We conducted the experiments on the \textit{All the News 2.0} dataset~\citep{thompsonAllNews202020},
a large-scale corpus of approximately 2.7 million news articles from
27 U.S. publications spanning the years 2016 to 2020.
We select this dataset
because it was not included in the original GPT-2 training corpus, ensuring that
our experiments evaluate dataset-scaling behavior without overlap with the data
used to pretrain the embedding representations. The dataset was accessed
via a publicly available Hugging Face distribution~\citep{thompsonAllNews202020}.
 
 The deduplicated and cleaned corpus is then tokenized using the standard GPT-2 tokenizer. GPT-2 operates with a fixed maximum context length of 1024 tokens. Since short articles introduce substantial padding, we analyze
padding utilization under different \emph{Minimum Article Length (MAL)} thresholds. Without length filtering, padding accounts for approximately 46\% of all processed tokens, and decreases monotonically with increasing MAL, dropping to 36.7\%, 32.4\%, and 28.4\% for thresholds of 256, 384, and 512 tokens, respectively. The padding ratio denotes the fraction of all processed tokens that correspond to padding after chunking articles into fixed-length sequences.


Based on this analysis, we adopt a MAL of 500 tokens for all
experiments, reducing the padding ratio to approximately 28\% while preserving
a sufficiently large and diverse corpus. All remaining sequences are padded or
truncated to 1024 tokens, with padded positions masked during self-attention and
excluded from loss and accuracy computation. We use a fixed training–validation partition, consisting of 131{,}072 training sequences and a held-out validation set of 20{,}000 sequences. The validation set is kept fixed and evaluated in full for all experiments, independent of training subset size. After
preprocessing, padding tokens account for approximately 28\% of all tokens
across both splits, a ratio that remains stable across subset sizes and random
samplings. Dataset statistics for the training and validation splits are
summarized in the Appendix \ref{app:ATN-stats}.

\subsection{Subset Construction}

To study dataset scaling in a controlled manner, we construct training
subsets by selecting a fixed number of sequences from the cleaned full
training set. Let $\mathcal{D} = \{x_1, \ldots, x_N\}$ denote the full training set, where each
$x_i$ is a token sequence of fixed length 1024. We define a sequence of nested subsets
$\{\mathcal{D}_k\}_{k=0}^{K}$, where the index $k$ denotes the subset
level, with cardinalities
\begin{equation}
|\mathcal{D}_k| = 2^k,\quad k \in \{7, \dots, 17\}.
\label{eq:subset-def}
\end{equation}

To ensure that subsets are nested $\mathcal{D}_k \subset \mathcal{D}_{k+1}$, we first generate a single random permutation of the full dataset using a fixed random seed. Each subset $\mathcal{D}_k$ is then defined as the first $2^k$ sequences of this permutation. This guarantees that larger subsets extend smaller ones without introducing additional sampling variability.

\subsection{Subset Analysis}

Beyond subset size alone, we analyze qualitative properties of the constructed subsets to ensure that observed performance differences can be attributed primarily to dataset size rather than systematic biases introduced by subset selection. In particular, we distinguish between \textit{(1) Distributional Similarity} to the full dataset and \textit{(2) Robustness to Random Subset Sampling}.

\subsubsection{Distributional Similarity}
We assess the extent to which randomly sampled training subsets approximate the
token distribution of the full corpus as subset size increases. While
distributional convergence is expected for sufficiently large random subsets,
our goal is to quantify the \emph{rate and scale} at which this convergence occurs
in our experimental setup. We compare token probability distributions of subsets of increasing size
to the reference distribution derived from the full dataset. For each subset, token frequencies are converted into probability distributions. These distributions are then compared using Jensen--Shannon (JS) divergence~\citep{linDivergenceMeasuresBased1991}, a bounded and symmetric measure of distributional similarity. Given two discrete probability distributions $p$
(subset distribution) and $q$ (full dataset distribution), JS divergence is
defined as

\begin{equation}
\mathrm{JS}(p \,\|\, q) =
\frac{1}{2}\mathrm{KL}(p \,\|\, m) +
\frac{1}{2}\mathrm{KL}(q \,\|\, m), 
\quad \text{where } m = \frac{1}{2}(p + q).
\label{eq:JS-div}
\end{equation}

\begin{figure*}[b]
  \centering
  \begin{minipage}[c]{0.48\textwidth} 
    \centering
    \includegraphics[height=5.5cm, width=\linewidth, keepaspectratio]{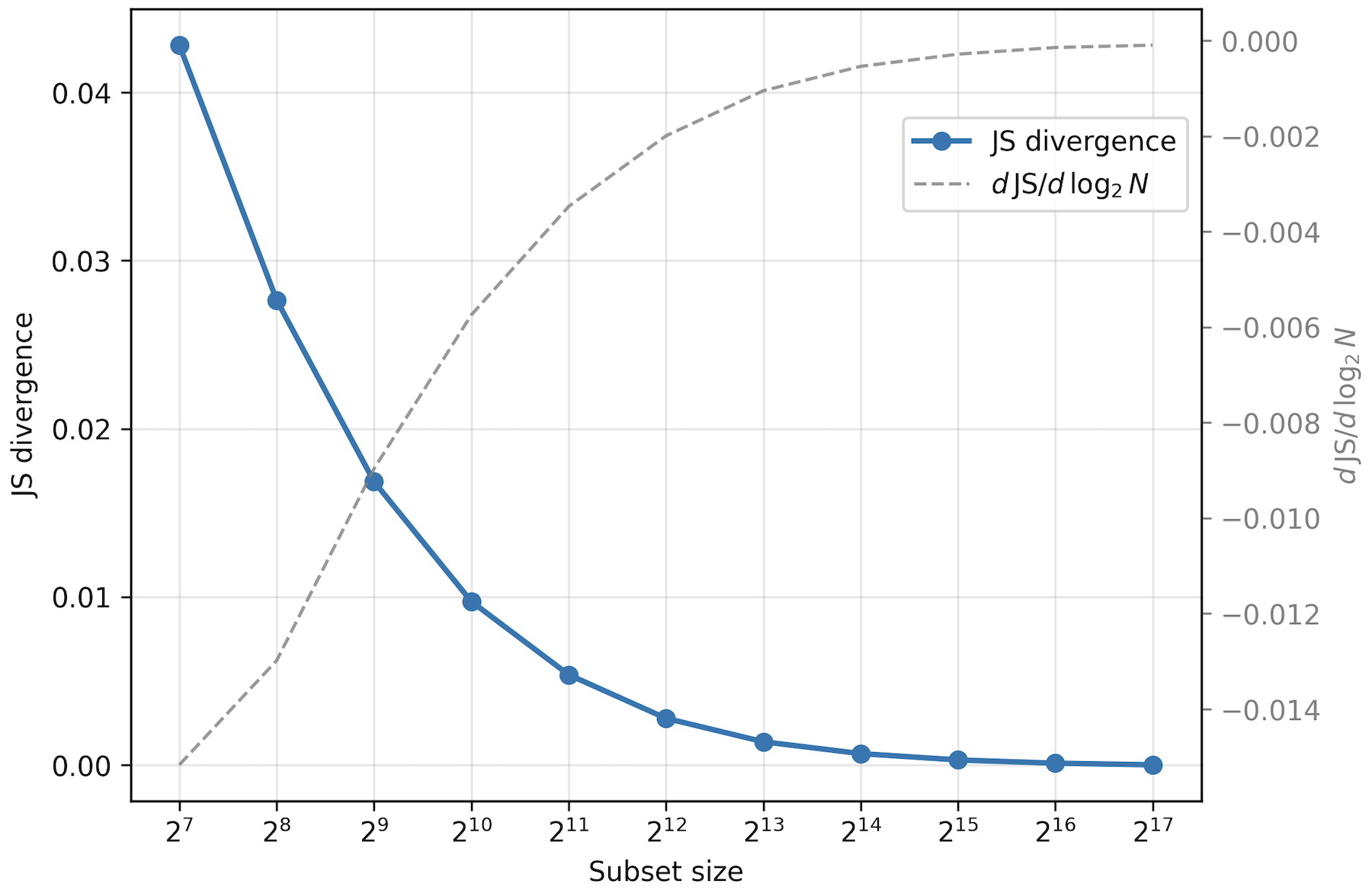}
  \end{minipage}
  \hfill
  \begin{minipage}[c]{0.48\textwidth}
    \centering
    \includegraphics[height=5.5cm, width=\linewidth, keepaspectratio]{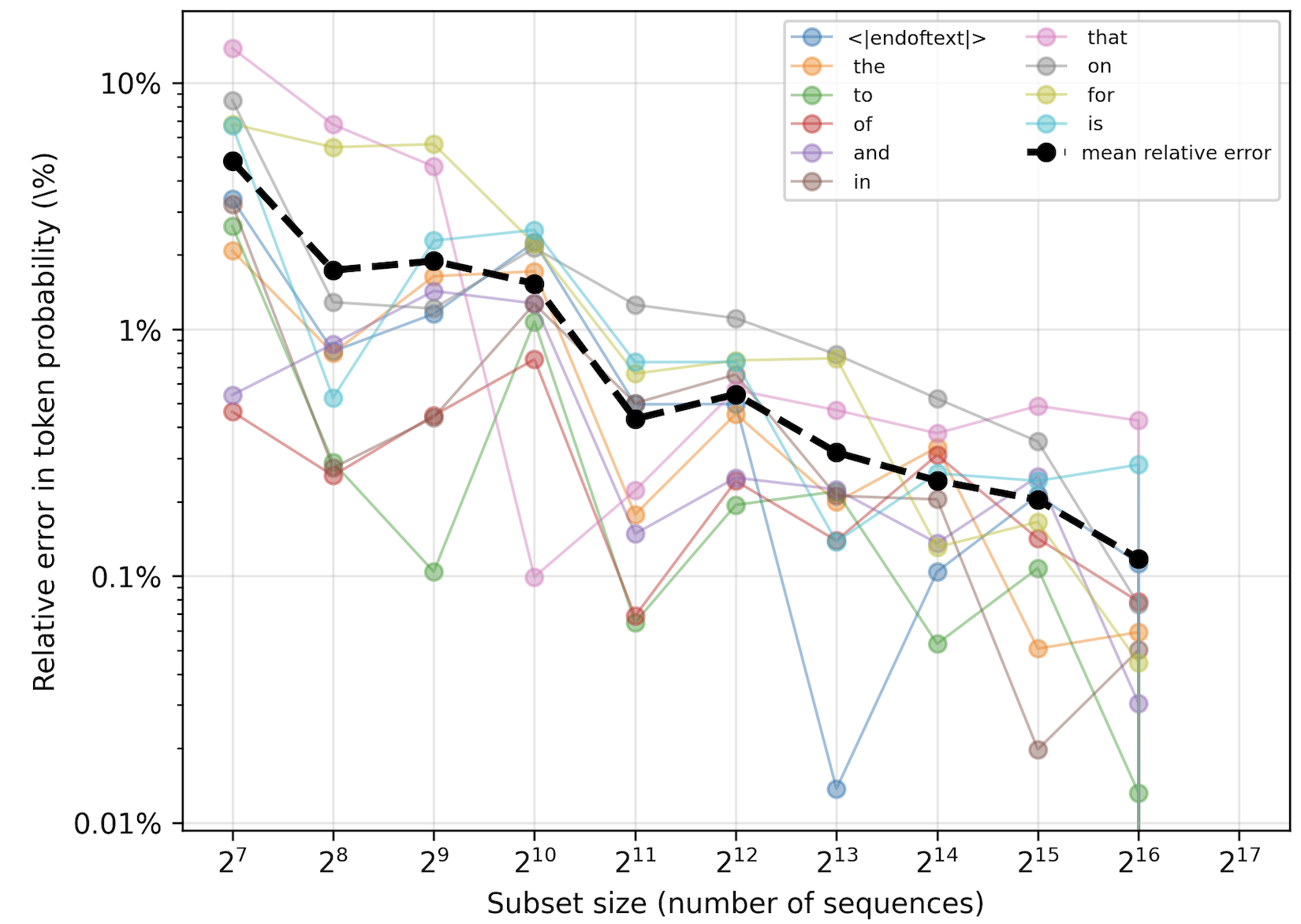}
  \end{minipage}
  \vspace{0.5em}
  \begin{minipage}[t]{0.48\textwidth}
    \caption{Jensen--Shannon divergence between token distributions of training subsets and the full dataset. The dashed curve shows the first derivative.}
    \label{fig:js_divergence_single}
  \end{minipage}
  \hfill
  \begin{minipage}[t]{0.48\textwidth}
    \caption{Convergence of relative token probability error for selected high-frequency tokens. The relative error is defined as $100 \cdot \lvert p_{\mathrm{subset}} - p_{\mathrm{ref}} \rvert / p_{\mathrm{ref}}$.}
    \label{fig:token-distribution}
  \end{minipage}
  
\end{figure*}

JS divergence quantifies the information-theoretic deviation of each
distribution from their shared middle distribution $m$. Here,
$\mathrm{KL}(p \,\|\, m) = \sum_i p_i \log \frac{p_i}{m_i}$ denotes the
Kullback--Leibler divergence. JS divergence values close to zero indicate that the subset and full
dataset token distributions are nearly identical. As shown in Fig.~\ref{fig:js_divergence_single}, JS divergence between subset and
reference token distributions decreases rapidly with increasing subset size,
indicating that moderate and large subsets closely approximate the statistical
structure of the full dataset. The same plot additionally shows the first derivative of JS divergence with respect to $\log_2 N$, which quantifies the rate at which distributional mismatch diminishes as more data is added. Small subsets ($2^{7}$ --$2^{8}$  sequences) exhibit
measurable distributional deviation, while for subset sizes of $2^{12}$  sequences and
above, JS divergence falls below 0.003, rendering subsets effectively
indistinguishable from the full dataset. 

In addition to distribution-level metrics, we analyze per-token relative
probability errors to study convergence behavior for frequent tokens (Fig. \ref{fig:token-distribution}). Relative errors between subset-based token probability estimates and the
reference distribution are shown for selected high-frequency tokens and their mean. Taken together, Figs.~\ref{fig:js_divergence_single} and~\ref{fig:token-distribution} aim to show \emph{scale and rate} at which
distributional convergence is reached in practice. The rapid decay of JS
divergence shows that already moderate-sized subsets recover the global token
distribution of the full dataset. Beyond this regime, the remaining differences
between subsets are small, supporting the interpretation that observed
differences in model performance are driven primarily by data quantity.

\subsubsection{Robustness to Random Subset Sampling}

To further validate these findings, we repeat training experiments across multiple random seeds for selected subset sizes. As shown in Appendix~\ref{app:seed_robustness}, variability across random seeds
decreases rapidly with increasing subset size, confirming that training outcomes are robust to sampling variability. This empirical stability justifies the use of a fixed random seed: for moderate
and large subsets, independently sampled subsets exhibit nearly identical token
frequency distributions, padding ratios and low standard deviation across seeds. Consequently, differences in model
performance for larger subsets can be attributed to data volume rather than
sampling-induced distributional bias.

\section{Experimental Setup}\label{sec:experimental}

\subsection{Training Setup}
Each training run is performed on a single Tesla V100-SXM3-32GB GPU using PyTorch (CUDA); the optimizer, learning rate, batch
size, model architecture, and training schedule are held constant across all
experiments- We train all models using the AdamW optimizer with a learning rate of
$3\times10^{-4}$ and a fixed batch size of 16 for 150 epochs. Dropout is disabled, and no learning rate scheduling or early stopping is applied. A single random
seed is used for all main experiments. Each training sequence consists of 1024 tokens. With a batch size of 16, each
optimization step processes $16 \times 1024 = 16{,}384$ tokens. Data subsets
are defined by the number of available sequences and therefore directly determine
the number of optimization steps per epoch according to \(\mathrm{\text{\emph{steps per epoch}} = \frac{\text{subset size}}{16}}\).

Prior analyses of fully trainable Transformer models indicate that compute-optimal
training is typically achieved when the number of training tokens
exceeds the number of model parameters by at least an order of magnitude~\citep{hoffmannTrainingComputeOptimalLarge2022a}. Importantly, this commonly cited heuristic was derived under the
assumption of full-model training, including embedding and output layers. The heuristic corresponds
to approximately $47$\,M training tokens for our architecture with $2.36\,\mathrm{M}$ trainable parameters. By
comparison, the full dataset used in our experiments processes
approximately $134$\,M tokens, while smaller subsets operate deep in the
data-limited regime. Table~\ref{tab:training_compute_merged} illustrates how increasing subset size systematically increases the effective
token-to-parameter ratio.

\subsection{Evaluation Protocol}

Models are evaluated on a fixed held-out validation set using masked
cross-entropy loss and masked token-level accuracy. We report the token-averaged cross-entropy $\mathcal{L}_{\text{token}}$ as the
primary validation loss, where averaging is performed over non-padding target
tokens, thereby weighting batches proportionally to their number of valid
tokens. Validation perplexity is computed as
$\exp(\mathcal{L}_{\text{token}})$. Token accuracy is defined as the fraction of
correct next-token predictions over non-padding target positions. Evaluation is performed prior to training (epoch~0), at every epoch for the
first 20 epochs, and subsequently every $N$ epochs ($N=10$), as well as at the
final epoch. All metrics are computed in evaluation mode without gradient tracking. We select the best checkpoint based on the minimum token-averaged validation loss $\mathcal{L}_{\text{token}}$.

\section{Results}
We study dataset scaling under two complementary experimental regimes to disentangle effects of data and compute. First, we analyze scaling under a fixed training schedule of 150 epochs, where increasing dataset size also increases compute budget (Section~\ref{sec:ATN}). Second, we consider a fixed-compute regime, where all configurations are trained with the same number of optimization steps (Section~\ref{sec:ATN-fixed-steps}).

\subsection{Subset Scaling and Cost-Performance Trade-off}
\label{sec:ATN}
\definecolor{c27}{HTML}{1F77B4} 
\definecolor{c28}{HTML}{FF7F0E} 
\definecolor{c29}{HTML}{2CA02C} 
\definecolor{c210}{HTML}{D62728} 
\definecolor{c211}{HTML}{9467BD} 
\definecolor{c212}{HTML}{8C564B} 
\definecolor{c213}{HTML}{E377C2} 
\definecolor{c214}{HTML}{7F7F7F} 
\definecolor{c215}{HTML}{BCBD22} 
\definecolor{c216}{HTML}{17BECF} 
\newcommand{\cdotc}[1]{\textcolor{#1}{\large$\bullet$}}

\begin{table}[ht]
  \centering
  \caption{Dataset statistics and training cost based on 2.4M trainable parameters.}
  \label{tab:training_compute_merged}
  \vspace{0.1cm}
  \setlength{\tabcolsep}{8pt} 
  \renewcommand{\arraystretch}{1.0} 
  \footnotesize

\tiny
\begin{tabular}{c@{}cccccc} 
    \multicolumn{2}{c}{\textbf{Subset Level}} &
    \textbf{\% Full Dataset} &
    \textbf{Size (seqs)} &
    \textbf{Tokens (in M)} &
    \textbf{Tokens/Param} &
    \textbf{Train Time (h:mm)} \\
    \toprule
    \cdotc{c27} & $2^{7}$   & 0.1  & 128       & 0.13   & 0.056 & 0:08 \\
    \cdotc{c28} & $2^{8}$   & 0.2  & 256       & 0.26   & 0.111 & 0:15 \\
    \cdotc{c29} & $2^{9}$   & 0.3  & 512       & 0.52   & 0.222 & 0:31 \\
    \cdotc{c210}& $2^{10}$  & 0.7  & 1{,}024   & 1.05   & 0.444 & 1:01 \\
    \cdotc{c211}& $2^{11}$  & 1.4  & 2{,}048   & 2.10   & 0.888 & 2:02 \\
    \cdotc{c212}& $2^{12}$  & 2.7  & 4{,}096   & 4.19   & 1.776 & 3:44 \\
    \cdotc{c213}& $2^{13}$  & 5.5  & 8{,}192   & 8.39   & 3.553 & 8:08 \\
    \cdotc{c214}& $2^{14}$  & 10.9 & 16{,}384  & 16.78  & 7.105 & 16:01 \\
    \cdotc{c215}& $2^{15}$  & 21.8 & 32{,}768  & 33.55  & 14.21 & 32:32 \\
    \cdotc{c216}& $2^{16}$  & 43.7 & 65{,}536  & 67.11  & 28.43 & 60:51 \\
    \cdotc{black}& $2^{17}$ & 100  & 131{,}072 & 134.22 & 56.86 & 123:55 \\
\end{tabular}
\normalsize

  \vspace{1.5em} 


  \includegraphics[width=1.0\textwidth]{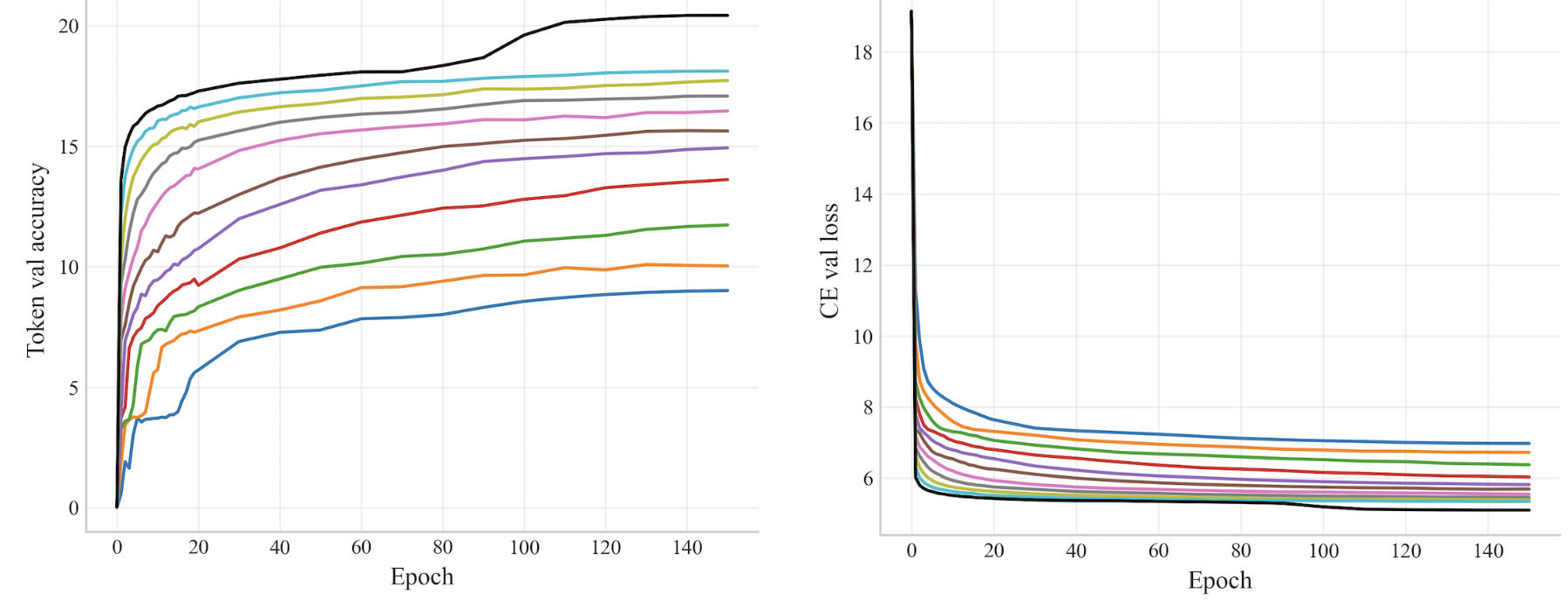} 
  \captionof{figure}{Training dynamics across dataset subset sizes}
  \label{fig:val_acc_epoch}
\end{table}

All results refer to the \textit{attention-only} architecture described in Section~\ref{sec:architecture} and training setup described in Section ~\ref{sec:experimental}. Fig.~\ref{fig:val_acc_epoch} shows validation token-level accuracy and CE-loss as a
function of subset size. These results quantify the cost–performance trade-off under a fixed training schedule, where increasing the dataset size also increases the total number of optimization steps. The reported gains reflect the joint effect of scaling both data and compute, where we observe clear diminishing returns. 
Early in the scaling regime, increasing subset size yields substantial accuracy gains. However, these gains quickly saturate, beyond moderate subset sizes, additional data provides only limited improvements despite a near-linear increase in training cost. This indicates that most achievable performance is captured early, while further scaling primarily increases compute without proportional benefit.



Table~\ref{tab:tma_tradeoff} reports the estimated dataset size and training cost
required to reach a given fraction of full-data accuracy. Concretely, reaching roughly 90\% of
the full-data accuracy requires only about 30\% of the training data. While the results are specific to the architecture and dataset used in our experiments, they provide practical guidance for reasoning about cost-performance trade-offs under fixed-dataset scaling and training schedule. 

\begin{table}[H]
\centering
\caption{Cost--performance trade-offs under fixed training schedule}
\vspace{0.2cm}
\setlength{\tabcolsep}{4pt}
\renewcommand{\arraystretch}{1.1}
\scriptsize
\begin{tabular}{c c c c c}
\hline
\textbf{Fraction of full-data accuracy} &
\textbf{Subset} &
\textbf{Data (\%)} &
\textbf{Train Time (h)} &
\textbf{Cost (\%)} \\
\hline
80\%  & $2^{13}$ & 7   & 8   & 7   \\
90\%  & $2^{15}$ & 29  & 32  & 26  \\
95\%  & $2^{16}$ & 56  & 60  & 49  \\
100\% & $2^{17}$ & 100 & 123 & 100 \\
\hline
\end{tabular}
\label{tab:tma_tradeoff}
\end{table}
\normalsize

This training setup deviates from the compute-optimal regime assumed in prior work, where models are trained until convergence. Since all subset sizes are trained for a fixed number of epochs, larger datasets receive fewer optimization steps and are comparatively undertrained. 

\subsection{Subset Scaling under fixed number of Optimization Steps}
\label{sec:ATN-fixed-steps}
To disentangle effects of data and compute, we analyse scaling under a fixed training budget (Fig.~\ref{fig:ATN-acc}, CE-Loss results in Appendix ~\ref{app:ATN-loss}). 
We observe the same pattern where performance initially improves with increasing dataset size but quickly saturates, and the largest subsets no longer provide consistent improvements under fixed compute, indicating that additional data alone is insufficient without a corresponding increase in training budget. To assess whether the observed behavior generalizes beyond a single dataset, we repeat the subset-scaling experiment on \textit{WikiText-103} dataset~\citep{merityPointerSentinelMixture2016}(see Appendix~\ref{app:wikitext-loss}), observing similar scaling dynamics. These results show that the observed diminishing returns reflect a broader interaction between dataset size and optimization steps. In particular, they highlight that the practical benefit of additional data depends critically on the available training budget.
\begin{figure}[htb]
    \centering
    \includegraphics[
    width=0.80\linewidth,
    trim={0.8cm 0.2cm 1.4cm 3.0cm},
    clip
]{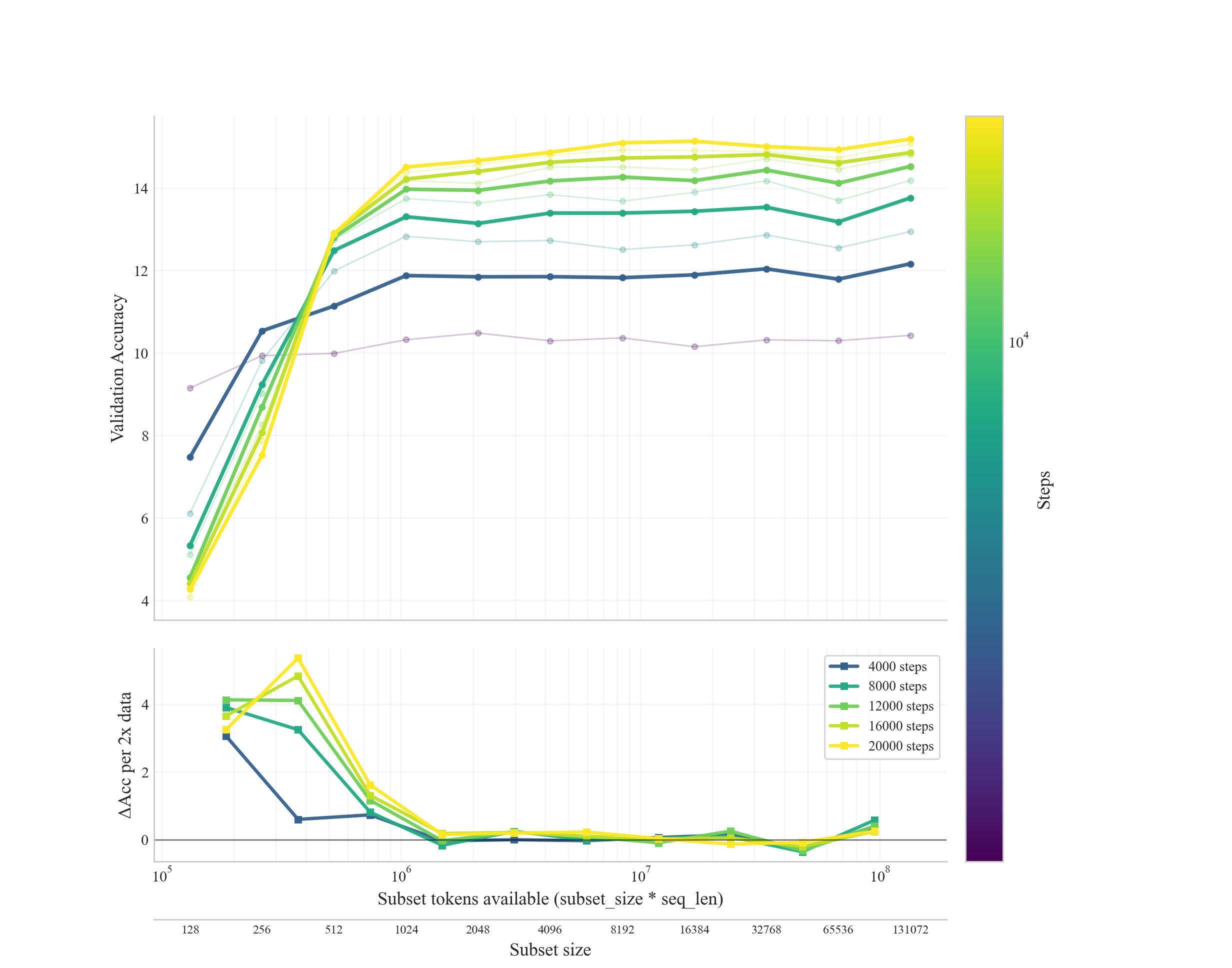}
    \caption{Validation accuracy on \textit{AllTheNews2.0} dataset vs. subset tokens for fixed training budget}
    \label{fig:ATN-acc}
\end{figure}

\subsection{Scaling-Law Analysis of Performance Limits}
To place the experimental setting of our \textit{attention-only} transformer in context, we relate our results on \textit{AllTheNews2.0} (Fig. ~\ref{fig:val_acc_epoch}) to the empirical scaling laws for language modeling
proposed by ~\citet{kaplanScalingLawsNeural2020a}. The scaling-law prediction provides a
reference for assessing whether, at a given model and dataset scale, performance
is primarily limited by data availability or by model capacity. Their analysis
models the test loss of a Transformer language model as a joint function of model
size $N$ and dataset size $D$:
\begin{equation}
L(N,D)=\left[\left(\frac{N_c}{N}\right)^{\alpha_N/\alpha_D}
+\left(\frac{D_c}{D}\right)\right]^{\alpha_D}
\label{eq:kaplan_scaling}
\end{equation}
where $\alpha_N$, $\alpha_D$, $N_c$, and $D_c$ are empirically fitted constants. Importantly, the scaling-law formulation assumes models are trained close to convergence under compute-optimal conditions. In contrast, our experiments use a fixed training schedule and do not ensure convergence for all subset sizes. Therefore, the following comparison should be interpreted as a qualitative reference for regime identification rather than a precise quantitative prediction.

Using the reported joint-fit constants, we evaluate the predicted loss for our
experimental setting:
\begin{align}
\alpha_N &= 0.076, &
\alpha_D &= 0.103, \nonumber \\
N_c &= 6.4 \times 10^{13}, &
D_c &= 1.8 \times 10^{13}.
\end{align}
Our model contains approximately $N=4.0\times10^{7}$ (40 million) parameters, including all parameters except token and positional embedding, and is
trained on $D\approx1.34\times10^{8}$ tokens (134 million). Substituting these values into \eqref{eq:kaplan_scaling} yields a predicted
test loss of
\begin{equation}
L(N,D)\approx 3.46 \ \text{nats/token.}
\label{pred-join-loss}
\end{equation}
The scaling law further allows us to estimate the asymptotic limits imposed by
data and model capacity.


In the limit of infinite data size,
\begin{equation}
L(N,\infty)=\left(\frac{N_c}{N}\right)^{\alpha_N}\approx 2.96,
\label{inf-data-limit}
\end{equation}
while in the limit of infinite model capacity,
\begin{equation}
L(\infty,D)=\left(\frac{D_c}{D}\right)^{\alpha_D}\approx 3.37.
\label{inf-model-limit}
\end{equation}

We compare the predicted joint loss in \eqref{pred-join-loss} to the two asymptotic limits. Since it is much closer to the infinite-model limit \eqref{inf-model-limit} than to the infinite-data limit \eqref{inf-data-limit}, performance is primarily constrained by data rather than model capacity.
Even with substantially
larger model capacity, the scaling law predicts only marginal loss improvements unless additional data is available. Since the predicted joint loss is already
near the infinite-model limit, increasing model capacity would yield only minor
improvements unless additional data is provided. This suggests that the
diminishing returns observed in our experiments are not caused by insufficient
model capacity or architectural constraints, but reflect a fundamental
data-limited regime.

This interpretation is further supported by a direct comparison between the empirical scaling behavior and the Kaplan prediction across all subset sizes (see Appendix~\ref{fig:kaplan_vs_actual}). For each subset, we evaluate \eqref{eq:kaplan_scaling} using the corresponding number of training tokens per subset size and our model size. While confirming that additional data remains beneficial and follows the expected power-law trend, the empirical loss saturates substantially earlier. We attribute this deviation primarily to the difference in training regimes: under fixed-epoch training, larger subsets are not trained to convergence, which leads to stronger apparent diminishing returns.

These findings have two implications for our experimental setup:

\begin{enumerate}
    \item They validate the use of a reduced, \textit{attention-only} architecture. Since performance is primarily constrained by data, our adopted architectural reductions
do not qualitatively alter the observed scaling behavior.
    \item They support subset-based training as an effective pretraining strategy for
small models, as performance saturates rapidly with increasing dataset size and
moderate subsets already capture most of the achievable performance. 
\end{enumerate}

Taken together, these results identify subset-based pretraining with early stopping
as principled strategies,
particularly for small or capacity-limited models. Consistent with scaling-law
predictions, the proximity of the predicted joint loss to the infinite-model
limit indicates that performance in our experiments is primarily data-limited
rather than model-limited.

\section{Conclusion}
We investigated whether the subset-scaling behavior observed for industrial-scale LLMs persists in the tiny-model regime. Using a controlled experimental setup, we trained a fixed-capacity \textit{attention-only} decoder aligned with the 117M GPT-2 design, in which token embeddings and the output projection are frozen and MLP sublayers are removed. This configuration isolates learning to the self-attention mechanism, enabling systematic analysis of dataset-size effects without architectural confounds. Our results suggest that the diminishing returns observed in our experiments are not caused by insufficient model capacity, but arise from a combination of data-scaling effects and limitations imposed by the finite training budget.

Across power-of-two training subsets, validation accuracy improves smoothly with increasing data and exhibits clear diminishing returns. Moderate subsets recover a large fraction of full-data performance at substantially reduced training cost, demonstrating that scaling behavior observed at industrial scale persists even for tiny decoder models. Robustness analyses further show that randomly sampled subsets are statistically representative of the full corpus and that performance trends are stable across seeds, indicating that observed gains are driven primarily by data volume rather than sampling artifacts.

Taken together, these results apply large-scale scaling-law insights to a practical, constrained setting. Rather than optimizing for the final percentage points of accuracy, our findings highlight subset-based pretraining with early stopping as a principled and efficient strategy for base model training. Tiny \textit{attention-only} decoders thus provide a valuable experimental substrate for studying scaling behavior and for rapid, economical pretraining in small research labs.

\newpage
\bibliography{IMP-ICLR2026}
\bibliographystyle{iclr2026_conference}

\newpage
\appendix
\section{Dataset Statistics \textit{AllTheNews2.0}}\label{app:ATN-stats}
\begin{table}[ht]
\centering
\caption{Training and validation dataset statistics of articles after cleaning ($>500$ tokens)}
\vspace{0.2cm}
\label{tab:dataset_stats}
\setlength{\tabcolsep}{10pt} 
\renewcommand{\arraystretch}{1.1}

\begin{tabular}{lrr} 
\textbf{Statistic} & \textbf{Training Dataset} & \textbf{Validation Dataset} \\
\midrule
Number of Sequences & 131{,}072 & 20{,}000 \\
Sequence Length & 1024 tokens & 1024 tokens \\
Total Token Count & 76{,}800{,}000 & 10{,}240{,}000 \\
Padding Tokens & 21{,}504{,}000 (28\%) & 2{,}867{,}200 (28\%) \\
\end{tabular}
\end{table}

\section{Robustness to Random Subset Sampling}\label{app:seed_robustness}
\begin{table}[ht]
\centering
\caption{Robustness to random subset sampling across seeds. Mean $\pm$ standard deviation over five seeds at epoch~1.}
\vspace{0.2cm}
\setlength{\tabcolsep}{4pt} 
\renewcommand{\arraystretch}{1.1}
\begin{tabular}{
    c 
    S[table-format=2.2(2), separate-uncertainty=true] 
    S[table-format=5.0(4), separate-uncertainty=true] 
    S[table-format=2.2(2), separate-uncertainty=true]
    }
{\textbf{Subset Level}} & {\textbf{Val. Loss}} & {\textbf{Val. Perplexity}} & {\textbf{Val. Accuracy (\%)}} \\
\midrule
$2^{7}$  & 11.24(0.07) & 76320(4973) & 0.40(0.12) \\
$2^{8}$  &  9.71(0.10) & 16453(1588) & 2.12(0.41) \\
$2^{9}$  &  8.68(0.05) &  5911( 268) & 3.13(0.21) \\
$2^{10}$ &  8.21(0.02) &  3683(  80) & 4.00(0.50) \\
$2^{11}$ &  7.70(0.08) &  2203( 178) & 5.74(1.20) \\
$2^{12}$ &  7.36(0.04) &  1574(  63) & 7.13(0.27) \\
$2^{13}$ &  7.20(0.06) &  1336(  86) & 7.80(0.31) \\
$2^{14}$ &  6.88(0.05) &   974(  47) & 9.18(0.28) \\
$2^{15}$ &  6.61(0.05) &   746(  35) & 10.37(0.19) \\
$2^{16}$ &  6.34(0.03) &   567(  16) & 11.87(0.20) \\

\end{tabular}
\end{table}

\section{Kaplan Predictions vs. Empirical Scaling}\label{kaplan-empri-calculated}

\begin{figure}[htb]
    \centering
    \includegraphics[width=0.7\textwidth]{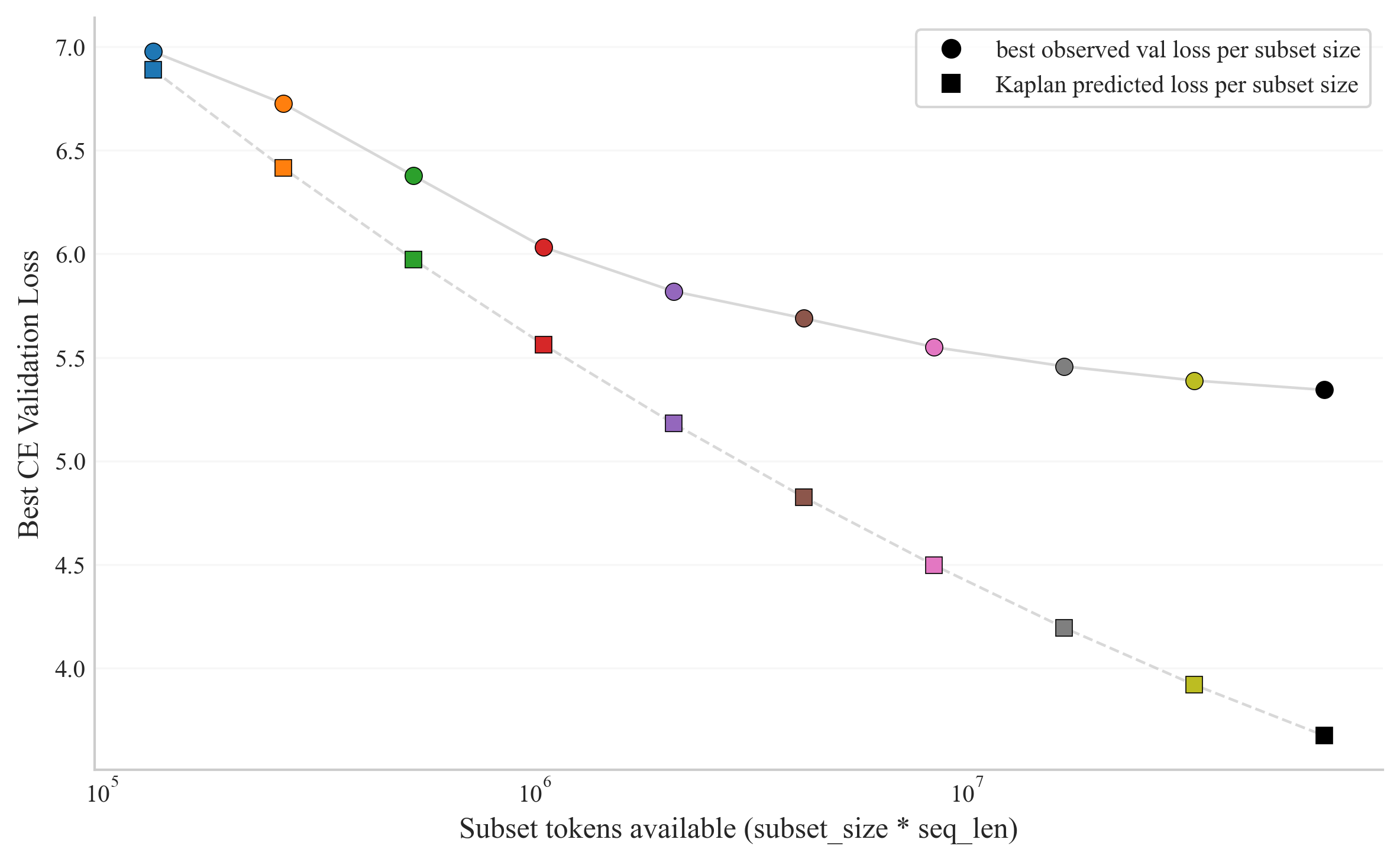} 
    \caption{Empirical best validation CE loss per subset size compared to the Kaplan scaling-law prediction on \textit{AllTheNews2.0}}
    \label{fig:kaplan_vs_actual}
\end{figure}

\newpage
\section{Subset Size Scaling Behavior on AllTheNews2.0 Dataset}\label{app:ATN-loss}
\begin{figure}[H]
    \centering
    \includegraphics[
    width=\linewidth,
    trim={0.2cm 0.2cm 0.5cm 3.0cm}, 
    clip
    ]{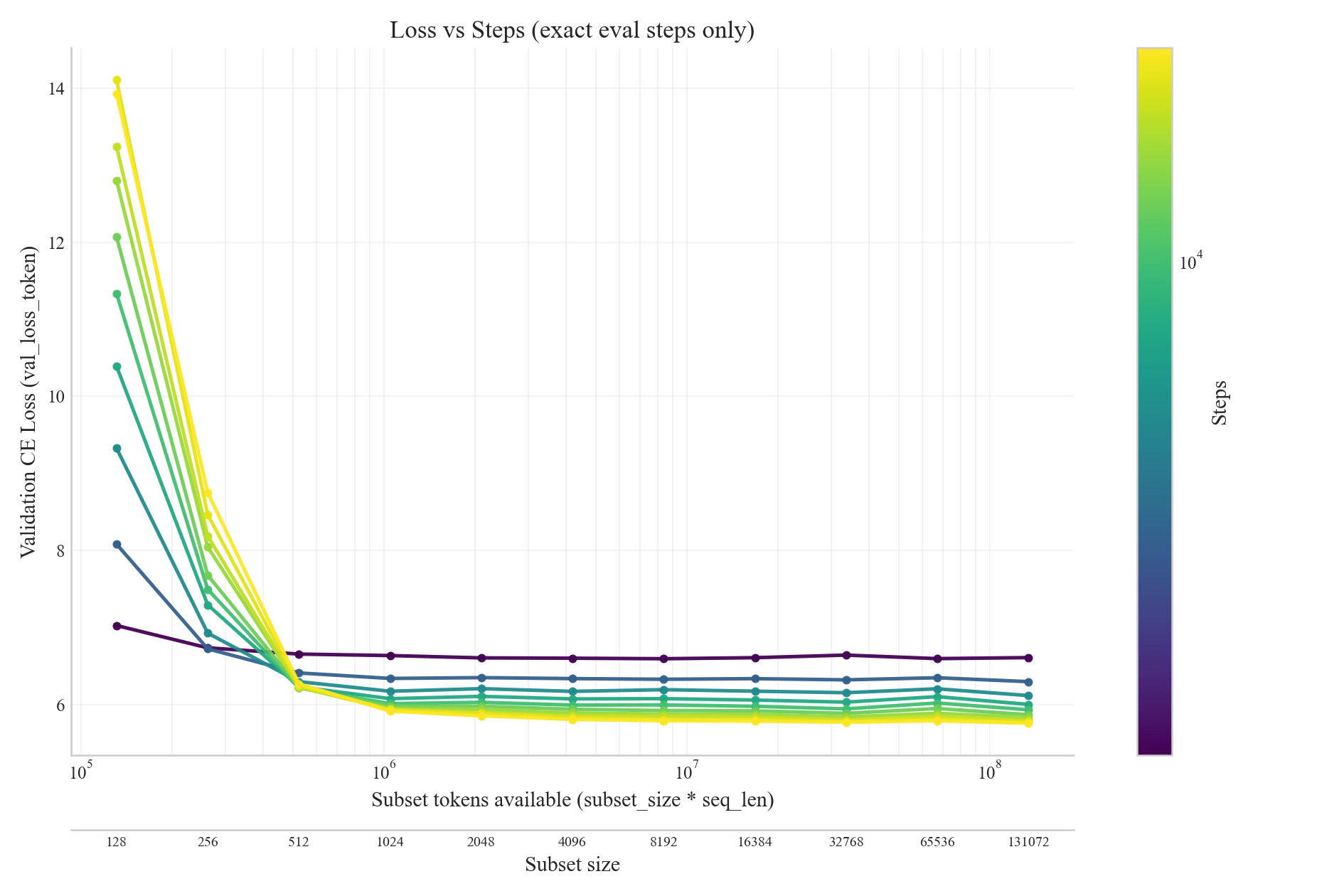}
    
    \caption{Cross-entropy loss on \textit{AllTheNews2.0} dataset under a fixed training budget.}
    \label{fig:ATN-loss-plot}
\end{figure}

\newpage
\section{Subset Size Scaling Behavior on WikiText-103 Dataset}\label{app:wikitext-loss}
\begin{figure}[H]
    \centering
    
    \includegraphics[
    width=\linewidth,
    trim={0.8cm 0.2cm 1.4cm 3.0cm},
    clip
    ]{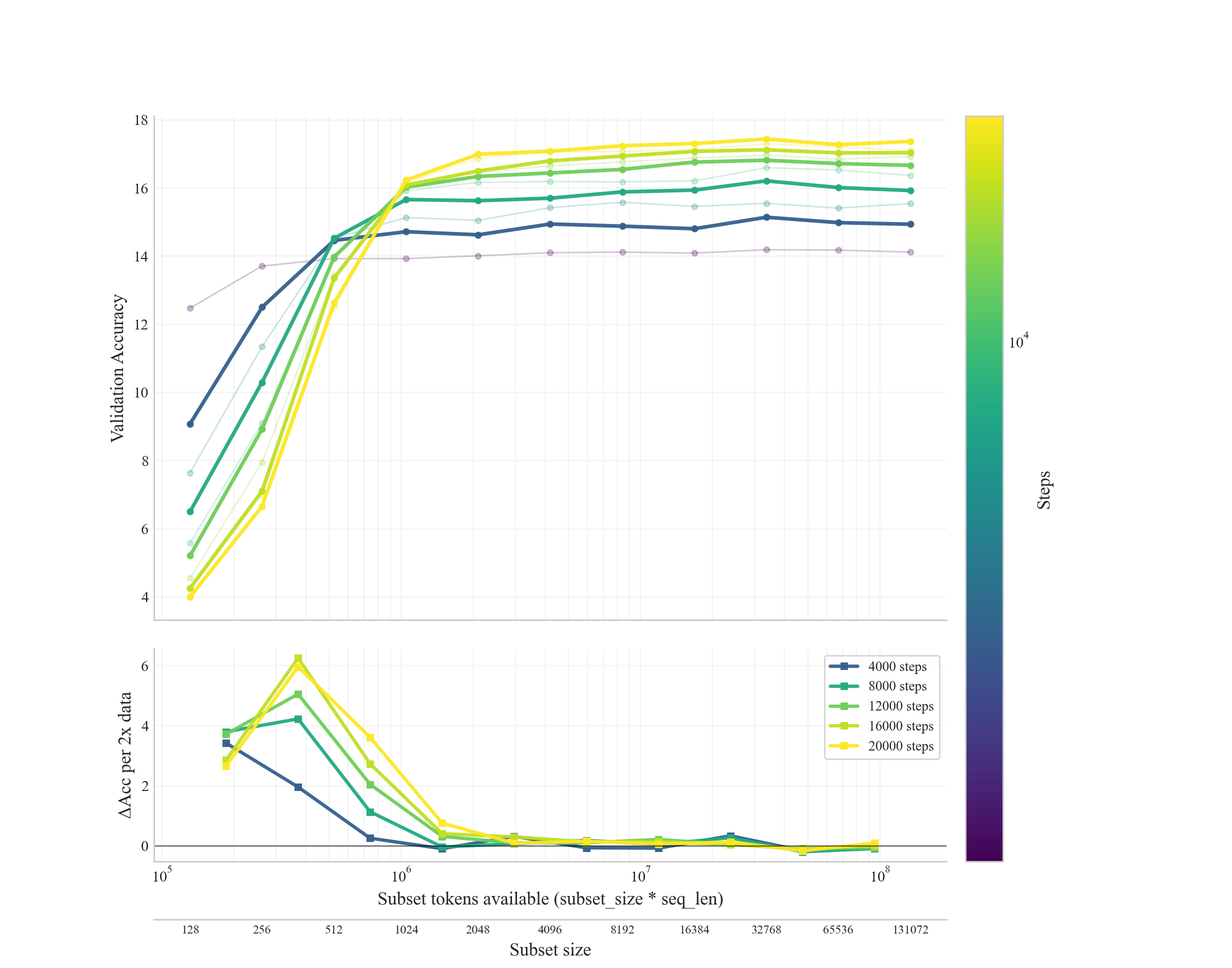}
    
    \vspace{0.5em}
    
    \includegraphics[
    width=\linewidth,
    trim={0.2cm 0.2cm 0.5cm 3.0cm}, 
    clip
    ]{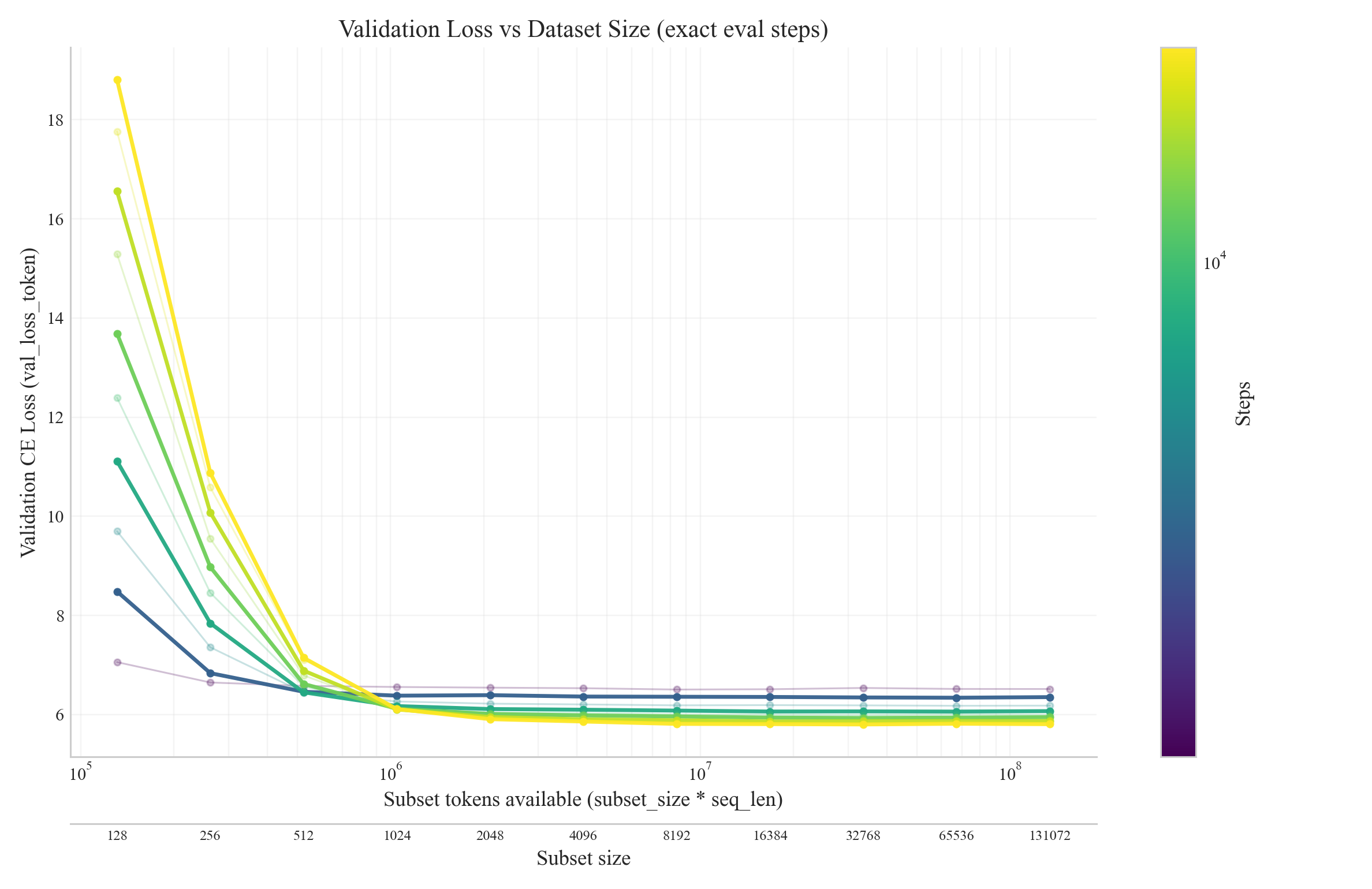}
    
    \caption{Validation accuracy (top) and cross-entropy loss (bottom) on \textit{WikiText-103} under a fixed training budget.}
    \label{fig:wikitext_combined}
\end{figure}

\end{document}